\documentclass[conference,10pt]{IEEEtran}
\usepackage{amsmath,graphicx,algorithm,algorithmic,xspace,amsfonts,amsthm,amssymb}
\usepackage{tikz}
\usepackage{pgfplots,multicol,blindtext}
\usepackage{cite}
\theoremstyle{definition}

\newtheorem{mylemma}{Lemma}

\newtheorem{proposition}{Proposition}

\begin{document}
	\title{A Connectedness Constraint for Learning Sparse Graphs}
	\author{
		\IEEEauthorblockN{Martin Sundin, Arun Venkitaraman, Magnus Jansson, Saikat Chatterjee}
		\IEEEauthorblockA{ACCESS Linnaeus Center, KTH Royal Institute of Technology, Sweden                 \\
			Email: masundi@kth.se, arunv@kth.se, janssonm@kth.se, sach@kth.se}
	}
%\ninept
%
\maketitle
\begin{abstract}
Graphs are naturally sparse objects that are used to study many problems involving networks, for example, distributed learning and graph signal processing. In some cases, the graph is not given, but must be learned from the problem and available data. Often it is desirable to learn sparse graphs.
However, making a graph highly sparse can split the graph into several disconnected components, leading to several separate networks. The main difficulty is that connectedness is often treated as a combinatorial property, making it hard to enforce in e.g. convex optimization problems.
In this article, we show how connectedness of undirected graphs can be formulated as an analytical property and can be enforced as a convex constraint. We especially show how the constraint relates to the distributed consensus problem and graph Laplacian learning.
Using simulated and real data, we perform experiments to learn sparse and connected graphs from data.
\end{abstract}
%
%\begin{keywords}
%Sparsity, graph signal processing, distributed inference, learning over networks.
%\end{keywords}
%

\IEEEpeerreviewmaketitle
\section{Introduction}
\label{sec:intro}

%\textcolor{red}{The paper is too long now. Limit is 4 pages + 1 page with references.}

Graphs are naturally sparse objects which describe relations between different data sources. Graphs are classically used in network problems, such as distributed estimation \cite{varshney1997distributed,sayed2013diffusion,boyd2011distributed,boyd2004fastest}. Other examples are webpage ranking \cite{page1999pagerank,langville2011google} and relations in social networks \cite{moura2014bigdata,sandryhaila2013discrete}. More recently, graphs have been used to generalize signal processing concepts, such as transforms, to signals defined on graphs \cite{moura2014bigdata,chen2015variation,venkitaramanhilbert, sandryhaila2014discrete,sandryhaila2013discrete,shuman2013emerging}. Often, the graph in question is sparse. For example, in social networks, there are many users and each user is only connected to few other users, resulting in a sparse graph.
%natural existence of sparsity.   
In many applications, the graph is not given a-priori but needs to be learned from data \cite{friedman2008sparse,mei2015signal, lin2012identification,gnecco2014sparse,learninglaplacian1,hegde2015nearly}. When learning sparse graphs, one challenge is that promoting sparsity may make the graph disconnected. 
Connectedness is hard to incorporate as it is often treated as a combinatorial property \cite{hegde2015nearly}. 
%The property of connectedness of a graph is combinatorial and hence it is hard to incorporate connectedness in learning optimization. 
In this article we show that connectedness is an analytical property that can be formulated as a convex constraint and subsequently be used efficiently in learning.

\subsection{Preliminaries and notations}

A graph $G = (\mathcal{V},\mathcal{E},\mathbf{A})$ is commonly defined as a set of nodes/vertices $\mathcal{V} = \{ 1,2,\dots, N \} = [N]$ and edges $\mathcal{E} \subset [N] \times [N]$ together with an adjacency matrix $\mathbf{A}\in\mathbb{R}^{N \times N}$. Two nodes $k\in V$ and $l\in V$ share an edge if $(k,l) \in \mathcal{E}$. The elements of the adjacency matrix describes the \emph{strength} of the connection between two nodes. The relation between $\mathcal{E}$ and $\mathbf{A}$ is:
%A graph $G = (V,\mathbf{A})$ can also be defined in terms of its adjacency matrix $\mathbf{A}\in\mathbb{R}^{N \times N}$ where two nodes $k\in V$ and $l\in V$ share an edge if and only if $A_{kl} \neq 0$. In the unweighted adjacency matrix all elements are either zero or one. The relation between the definitions is thus
$
(k,l) \in \mathcal{E} \Leftrightarrow A_{kl} \neq 0.
$
For undirected graphs, the adjacency matrix $\mathbf{A}$ is symmetric. 
 %and $(k,l) \in E \Leftrightarrow (l,k) \in E$. 
In this article we consider real symmetric adjacency matrices with non-negative components. 
 %undirected graphs.
%In distributed consensus, the matrix $\mathbf{A}$ contains the weights of the consensus algorithm. In the graphical LASSO, $\mathbf{A}$ is a regularized inverse covariance matrix and in GSP, the matrix $\mathbf{A}$ is the graph translation operator. The elements of $\mathbf{A}$ can be positive or negative. As the matrix $\mathbf{A}$ completely characterizes the graph $G$ we sometimes write $G = G(\mathbf{A})$(?).

%\subsection{Notation}

We denote the matrix determinant by $\mathrm{det}(\cdot)$, the matrix trace by $\mathrm{tr}(\cdot)$. We use $\succ$ ($\succeq$) to denote positive (semi) definiteness of symmetric matrices and $\geq$ to denote element-wise greater or equal. We use $\mathbf{1} = (1,\,1,\,1,\, ... \, ,1^\top \in\mathbb{R}^N$ to denote the vector consisting of only ones and $\mathbf{I}_N$ to denote the $N \times N$ identity matrix. The $k$'th largest eigenvalue of a matrix is denoted by $\lambda_k(\cdot)$ and the element-wise $\ell_1$-norm by $||\mathbf{A}||_1 = \sum_{k,l} |A_{kl}|$.% The uniform distribution on an interval $[a,b]$ is denoted by $\mathcal{U}[a,b]$.

\subsection{Problem statement}

Many graph learning problems can be posed as finding the adjacency matrix $\mathbf{A}$ that minimizes an objective function $g(\mathbf{A})$ under appropriate constraints. To make the adjacency matrix sparse, a sparsity promoting penalty function $s(\mathbf{A})$ is often introduced. The graph is thus obtained by solving the optimization problem:
\begin{align}
\label{eq:g_s_optimization}
\begin{aligned}
\min_{\mathbf{A}} &\,\, g(\mathbf{A}) + \eta s(\mathbf{A}), \,\, \mbox{subject to} \,\, constraints
\end{aligned}
\end{align}
where $\eta > 0$ is a regularization parameter and $\mathbf{A}$ is symmetric. A few examples of $g(\mathbf{A})$, $s(\mathbf{A})$ and $constraints$ are shown in Table~\ref{tab:graph_problems}. Signal Processing on Graphs (SPG) and the graphical LASSO are data-driven problems while the distributed consensus problem only uses the topology of an underlying graph. We note that all optimization problems in Table~\ref{tab:graph_problems} are convex.

\begin{table*}[t]
\small
\begin{center}
\renewcommand{\arraystretch}{1.2}
\begin{tabular}{|l|c|c|c|}
\hline
\textbf{Problem} & $g(\mathbf{A})$ &  $s(\mathbf{A})$ & \textbf{constraints}\\
\hline
Consensus \cite{boyd2004fastest,lin2012identification,gnecco2014sparse} & $ \mu(\mathbf{A}) = \max\{ \lambda_2(\mathbf{A}),-\lambda_N(\mathbf{A})\}$ & $||\mathbf{A}||_1 - \mathrm{tr}(\mathbf{A})$ & $\mathbf{A1=1}$, $\mathbf{A} \geq \mathbf{0}$, $A_{ij}=0$ for $(i,j) \notin \mathcal{E}$\\
Graphical LASSO \cite{friedman2008sparse} & $\mathrm{tr}(\hat{\mathbf{R}}\mathbf{A}) - \log \mathrm{det}(\mathbf{A})$ & $\|\mathbf{A}\|_1$& $\mathbf{A} \succeq \mathbf{0}$ \\
%LASSO & & & \\
SPG \cite{shuman2013emerging,learninglaplacian1} & $\mathrm{tr}(\mathbf{X} \mathbf{L} \mathbf{X}^\top)$ & $\|\mathbf{A}\|_1$ & $\mathbf{L} = \mathrm{diag}(\mathbf{A}\mathbf{1}) - \mathbf{A}$ \\
\hline
\end{tabular}
\end{center}
\vspace*{-0.cm}
\caption{Few examples of $g(\mathbf{A})$, $s(\mathbf{A})$ and $constraints$ that can be used in \eqref{eq:g_s_optimization}.}
\label{tab:graph_problems}
\end{table*}

Often, the learned graph becomes disconnected for large values of $\eta$. This means that the resulting graph describes two or more separate non-interacting systems. In this article, we address the issue of learning sparse connected graphs by formulating a convex constraint which preserves connectedness. Our contributions are as follows:
\begin{enumerate}
\item We analytically formulate connectedness in terms of a weighted Laplacian matrix.
\item For the distributed consensus problem we show that the graph splits into several components when the parameter $\eta$ exceeds some value. 
\end{enumerate}
We illustrate the validity of the proposed constraint through numerical simulations considering both synthetic data and real world temperature data.

% The constraint is applicable to all non-directed graph problems, however, for concreteness we especially consider learning graphs for graph signal processing and the distributed consensus problem.

\section{Preserving the connectedness of graphs}
\label{sec:connected}

The connectedness of a graph is usually described by the spectrum of the graph Laplacian matrix defined using the incidence matrix \cite{Chung}. As the Laplacian is not a continuous function of the adjacency matrix $\mathbf{A}$, preserving connectedness in terms of the Laplacian matrix leads to combinatorial optimization problems. For this reason we instead consider the weighted graph Laplacian matrix $\mathbf{L} \in \mathbb{R}^{N \times N}$ with elements
\begin{align*}
L_{kl} = \left\{ \begin{array}{ll}
\sum_{m\neq k} A_{km} & \text{, if } k = l,\\
-A_{kl} & \text{, if } k \neq l ,\\
\end{array}
\right.
\end{align*}
the matrix can thus be expressed as $\mathbf{L} = \mathrm{diag}(\mathbf{A}\mathbf{1}) - \mathbf{A}$. 
%To show that the weighted Laplacian is positive semi-definite 
We note that for $\mathbf{x}=[x_1,\cdots,x_N]^T$
\begin{align*}
\mathbf{x}^\top \mathbf{L} \mathbf{x} %&= \sum_{k,l} |A_{kl}| x_k^2 - \sum_{k,l} |A_{kl}|x_kx_l \\&
&= \sum_{k,l} A_{kl} \left( x_k^2 - x_kx_l \right) = \frac{1}{2} \sum_{k,l} A_{kl} \left( x_k^2 - 2x_kx_l + x_l^2\right)\\
&= \frac{1}{2}\sum_{k,l} A_{kl} (x_k - x_l)^2 = \sum_{k<l} A_{kl} (x_k - x_l)^2.
\end{align*}
The weighted Laplacian is thus positive semi-definite when $A_{ij} \geq 0$. The nullspace of the weighted Laplacian relates to the number of connected components through the following lemma.
%When the adjacency matrix $\mathbf{A}$ is non-negative, the weighted Laplacian is a linear function of $\mathbf{A}$. 
%The key insight is that the dimension of the null space of $\mathbf{L}$ equals the number of connected graph components. We formulate this as a proposition.
%The number of connected components in a graph is related to the eigenspectrum of its weighted Laplacian through the following lemma:  
\begin{mylemma}[Connected components]
\label{lemma:connected_nullspace}
The number of connected components of a graph with adjacency matrix $\mathbf{A}$ equals to the dimension of the null space of $\mathbf{L}$.
%, i.e.
%\begin{align*}
%\# \text{connected components of } G = \mathrm{dim}(\mathrm{Null}(\mathbf{L})).
%\end{align*}
\end{mylemma}

The proof of Lemma~\ref{lemma:connected_nullspace} is similar to the proof for the (unweighted) Laplacian in \cite{Chung} and is therefore not repeated here. Lemma~\ref{lemma:connected_nullspace} gives that a graph is connected if and only if the second smallest eigenvalue of $\mathbf{L}$ (also known as the Fiedler value) is nonzero. By noting that $\mathrm{span}\{\mathbf{1}\} \subset \mathrm{null}(\mathbf{L})$, we find the following proposition.

%$\mathrm{null}(\mathbf{L}) = \mathrm{span}\{\mathbf{1}\}$. This can be equivalently expressed as the condition that  Through this observation, we next propose the following proposition.%theorem which shows that the connectedness constraint is a convex constraint on the adjacency matrix $\mathbf{A}$:

%As the vector $\mathbf{1}$ always lie in the null space of $\mathbf{L}$, to ensure connectedness, we need to ensure that no other vectors lie in the null soace of $\mathbf{L}$. We formulate this as a theorem.

\begin{proposition}[Graph connectedess constraint]
\label{thm:laplace_connected}
A graph with adjacency matrix $\mathbf{A} \geq \mathbf{0}$ is connected if and only if
\begin{align}
\label{eq:connected_constraint}
\mathbf{L} + \frac{1}{N} \mathbf{11^\top} = \mathrm{diag}(\mathbf{A}\mathbf{1}) - \mathbf{A} + \frac{1}{N} \mathbf{11^\top} \succ \mathbf{0}.
\end{align}
%for some $\epsilon > 0$. 
\end{proposition}

\begin{proof}
%\textcolor{red}{Martin: Can you please add the proof here?}\\
%We show the proposition by showing that there exists a vector $\mathbf{x}$ such that $\mathbf{x}^\top(\mathbf{L} + \frac{1}{N} \mathbf{11^\top})\mathbf{x} = 0$ if and only if the graph is disconnected.

Let the graph be $G=(\mathcal{V},\mathcal{E},\mathbf{A})$. Assume first that there exists a vector $\mathbf{x} \neq \mathbf{0}$ such that
\begin{align*}
\mathbf{x}^\top \left(\mathbf{L} + \frac{1}{N} \mathbf{11^\top} \right)\mathbf{x} = \sum_{i > j} A_{ij} (x_i - x_j)^2 + \frac{1}{N}\left( \sum_{i=1}^N x_i \right)^2 ,
\end{align*}
is zero. This means that $\sum_{i=1}^N x_i = 0$ and $x_i = x_j$ for all $i$ and $j$ such that $A_{ij} > 0$. Hence, $x_j = x_i$ for all nodes $j\in V$ connected to a the node $i\in V$, the vector $\mathbf{x}$ is thus piecewise constant over the graph. However, the vector cannot be completely constant since $\sum_{i=1}^N x_i = 0$ and $\mathbf{x \neq 0}$ by assumption. The graph must therefore consist of at least two components.

Next, assume that the graph is disconnected. Then there exists two sets $C$ and $D$ such that $C \cup D = \mathcal{V}$, $C \cap D = \emptyset$ and $A_{ij} = 0$ for all $(i,j) \in C \times D$. Set the elements of $\mathbf{x}$ be $x_i = 1/|C|$ for $i \in C$ and $x_j = - 1/|D|$ for $j \in D$ where $|C|$ and $|D|$ denotes the number of elements in $C$ and $D$ respectively. We find that
\begin{align*}
\mathbf{x}^\top(\mathbf{L} + \frac{1}{N} \mathbf{11^\top})\mathbf{x} = %\sum_{i > j, i \in C, j \in D} |A_{ij}|\left(\frac{1}{n_C} + \frac{1}{n_D} \right)^2 \\
%+ 
\frac{1}{N}\left( |C| / |C| - |D| / |D| \right)^2 = 0.
\end{align*}
Hence \mbox{$\mathbf{L} + \frac{1}{N} \mathbf{11^\top} \nsucc \mathbf{0}$}.
\end{proof}
Proposition~\ref{thm:laplace_connected} immediately implies that the solution to the graph learning problem \eqref{eq:g_s_optimization} is connected when the constraint
\begin{align}
\mathrm{diag}(\mathbf{A}\mathbf{1}) - \mathbf{A} + \frac{1}{N} \mathbf{11^\top} \succeq \epsilon \mathbf{I}_N, \label{cvx_constraint}
\end{align}
is imposed for some $\epsilon > 0$ and $A_{ij} \geq 0$.

For the graphical Lasso problem \cite{friedman2008sparse}, the components of the adjacency matrix $\mathbf{A}$ are not necessarily non-negative. However, the sign pattern of the solution is the same as that of $\hat{\mathbf{R}}$ in the sense that $\hat{R}_{ij} A_{ij} \geq 0$ for all $(i,j)$ \cite{friedman2010applications, glasso_thresholding}. One can therefore replace the adjacency matrix in \eqref{cvx_constraint} by an adjacency matrix $\tilde{\mathbf{A}}$ with components $\tilde{A}_{ij} = \mathrm{sign}(\hat{R}_{ij}) A_{ij}$. The constraint can therefore also be used in problems with negative adjacency matrices provided that the sign pattern of $\mathbf{A}$ is known.

We next consider the application of the constraint to signal processing on graphs and the distributed consensus problem.

\section{Applications}

\subsection{The consensus problem}

In the distributed consensus problem \cite{boyd2004fastest,gnecco2014sparse,lin2012identification}, we iteratively compute the mean of a sequence $\{ x_i(0) \}_{i=1}^N$ as%through the rule
%are interested in distributively computing the mean value of a set of data $\{ x_i(0) \}_{i=1}^N$. The mean 
%is computed iteratively at each node through the consensus rule
\begin{align*}
x_i(t+1) = \sum_{j} A_{ij}x_j(t); \,\, t=0,1,2,\dots ,
\end{align*}
where $A_{ij} \geq 0$ and $x_j(t)$ denotes the value at the $j$'th node at the $t$'th iteration. The iterations converge to the mean as $t \to \infty$ when $\mathbf{A1 = 1}$ and $\mu(\mathbf{A}) = \max\{ \lambda_2(\mathbf{A}),-\lambda_N(\mathbf{A})\} < 1$ \cite{boyd2004fastest}. Smaller $\mu(\mathbf{A})$ leads to faster (worst case) convergence.

In some scenarios, for example when the number of communication links is limited, it is desirable to obtain a sparse graph \cite{gnecco2014sparse,lin2012identification}. 
This can be done by setting \cite{gnecco2014sparse}
%This can be achieved by using the sparsity promoting penalty \cite{gnecco2014sparse}
\begin{align*}
s(\mathbf{A})= \sum_{i \neq j} A_{ij} = N - \mathrm{tr}(\mathbf{A}) = \mathrm{tr}(\mathbf{I}_N - \mathbf{A}) = \mathrm{tr}(\mathbf{L}).
\end{align*}
The optimization problem \eqref{eq:g_s_optimization} then becomes \cite{gnecco2014sparse}:
\begin{align}
\label{eq:consensus_optimization}
\begin{aligned}
\min & \,\, \mu(\mathbf{A}) + \eta \mathrm{tr}(\mathbf{I}_N - \mathbf{A}),\\
\text{subject to} & \,\, \mathbf{A1 = 1}, \,\, \mathbf{A} \geq \mathbf{0},\,\, \mathbf{A^\top = A}\\
& \,\, A_{kl} = 0 \text{, for all } (k,l) \notin \mathcal{E}.
\end{aligned}
\end{align}
The optimization problem \eqref{eq:consensus_optimization} does not ensure that the graph is connected. In fact, we now prove that the solution to \eqref{eq:consensus_optimization} is guaranteed to be disconnected for certain values of $\eta$.% We state this in the following proposition.

\begin{proposition}[Consensus graph splitting with sparsity penalty]
\label{thm:consensus_split}
In the consensus problem \eqref{eq:consensus_optimization}, the graph with $N$ nodes and adjacency matrix $\mathbf{A}$ splits up into at least $k$ connected components when
\begin{align*}
\eta > 1 / (N - k+1), \,\, \text{for} \,\, k=1,2,\dots, N.
\end{align*}
%for $k=1,2,\dots, N$.
\end{proposition}
\begin{proof} We prove the proposition by induction over $k$. The graph consists of at least one connected component, so the proposition holds for $k=1$. Assume that the proposition holds for $k-1$. Since $\mathbf{L} \succeq \mathbf{0}$, the eigenvalues $\lambda_1 \geq \lambda_2 \geq \dots \geq  \lambda_N = 0$ of $\mathbf{L}$ are non-negative. The induction hypothesis gives that $\lambda_{N-k+2} = \lambda_{N-k+3} = \dots = \lambda_N = 0$. We find that
\begin{align*}
1 &= \mu(\mathbf{I}_N) \geq \mu(\mathbf{A}) + \eta \mathrm{tr}(\mathbf{I}_N-\mathbf{A}) = \mu(\mathbf{I}_N-\mathbf{L}) + \eta \mathrm{tr}(\mathbf{L}) \\
&\geq 1-\lambda_{N-k+1} + \eta (\lambda_{N-k+1} + \lambda_{N-k+2} + \dots + \lambda_1) \\
&\geq 1 + \lambda_{N-k+1} \left( \eta (N-k+1) -1 \right).
\end{align*}
So $\lambda_{N-k+1} \left( \eta (N-k+1) - 1\right) \leq 0$. Since $\lambda_{N-k+1} \geq 0$, we find that $\lambda_{N-k+1} = 0$ if $\eta > \frac{1}{N-k+1}$. Hence, for $\eta > \frac{1}{N-k+1}$, $\mathrm{dim} (\mathrm{null}(\mathbf{L})) \geq k$ and the graph consists of at least $k$ connected components.
\end{proof}

Proposition~\ref{thm:consensus_split} generalizes the result from \cite{gnecco2014sparse} which states that the graph becomes completely disconnected (consists of $N$ independent components) when $\eta > 1$. 

Next we examine the effect of imposing the connectedness constraint \eqref{cvx_constraint} on the consensus problem. 
%Since the mixing matrix $\mathbf{A}$ is positive and $\mathbf{A1 = 1}$, we have that
%The connectedness constraint gives us that
%\begin{align*}
%\mathbf{L} + \frac{1}{N}\mathbf{11^\top} = \mathbf{I}_N - \mathbf{A} + \frac{1}{N}\mathbf{11^\top}  \succeq \epsilon \mathbf{I}_N.
%\end{align*}
%For $\mathbf{L} = \mathbf{I}_N - \mathbf{A}$, 
The constraint can be rewritten as
\begin{align*}
\mathbf{A} - \frac{1}{N}\mathbf{11^\top} \preceq (1-\epsilon)\mathbf{I}_N .
\end{align*}
This gives us that
\begin{align*}
\lambda_2(\mathbf{A}) = \lambda_1 \left( \mathbf{A} - \frac{1}{N}\mathbf{11^\top} \right)  \leq  1-\epsilon.
\end{align*}
This shows that the graph is connected when $\lambda_2(\mathbf{A})  \leq \mu(\mathbf{A}) < 1$. Convergence of the consensus rule ($\mu(\mathbf{A}) < 1$) thus implies connectedness, whereas the reverse is not true. 
%Connectedness is not equivalent to convergence of the consensus rule. 
For example, when $\lambda_N(\mathbf{A}) = -1$ and $\lambda_2(\mathbf{A}) < 1$, the graph is connected but the consensus rule is not guaranteed to converge.

\subsection{Signal processing on graphs}

Signal processing on graphs (SPG) \cite{shuman2013emerging,sandryhaila2013discrete,sandryhaila2014discrete,moura2014bigdata} is an emerging field which deals with processing and analysis of data defined over graphs. Concepts such as sampling, filters and transforms have been generalized to graph signals \cite{shuman2013emerging,sandryhaila2014discrete,venkitaramanhilbert,jung_scalable,chen2015variation,shuman2016vertex}. In many problems the graph needs to be learned from data \cite{learninglaplacian1,7532684,mei2015signal}. The aim is then to find a sparse adjacency matrix that describes similarities in a given data set. For undirected graphs, a commonly used measure of graph signal smoothness is the Laplacian quadratic form
\begin{align*}
%l(\mathbf{x})= 
\mathbf{x^TLx}= \sum_{k<l} A_{kl}(x_k - x_l )^2,
\end{align*}
where $A_{kl} \geq 0$. 
%The Laplacian quadratic form is equal to the sum of the squared differences in the values of the graph signal across connected edges, weighted by the edge weight value. 
A graph signal $\mathbf{x}$ with a small $\mathbf{x^TLx}$ varies smoothly across the edges of the graph \cite{shuman2013emerging,chen2016representations,chen2015variation,learninglaplacian1}.
%Learning the Laplacian $\mathbf{L}$ can therefore be used to find similarities between different time series \cite{shuman2013emerging}. 
Given $N$ time series in a matrix $\mathbf{X} = [\mathbf{x}_1, \, \mathbf{x}_2, \, \dots ,\mathbf{x}_N] \in \mathbb{R}^{M \times N}$, one can find graphs describing the smoothness in the data by setting%minimizing the objective
% becomes%average of the Laplacian quadratic forms can be written as
% the goal is to find a sparse and connected graph with adjacency matrix $\mathbf{A}$ that describes the similarities between the time series.
% objective function $g(\mathbf{A})$ is then
\begin{align*}
g(\mathbf{A}) = \frac{1}{M} \sum_{t=1}^M \sum_{i>j} A_{ij}(X_{ti} - X_{tj})^2 = \frac{1}{M} \mathrm{tr}(\mathbf{X} \mathbf{L} \mathbf{X}^\top).
\end{align*}
Self-edges can be removed by setting $A_{ii} = 0$ for all $i \in \mathcal{V}$ and the rows can be normalized by setting $\mathbf{A1 = 1}$ to prevent the trivial solution $\mathbf{A = 0}$, giving that $ ||\mathbf{A}||_1 = N$. The connected graph learning problem thus becomes
\begin{align}
\label{eq:spg_learning_problem}
\begin{array}{ll}
\min\limits_{\mathbf{A}} & \frac{1}{M}\mathrm{tr}\left(\mathbf{X} \left(\mathbf{I}_N - \mathbf{A} \right) \mathbf{X^\top}\right)\\
\text{subject to} & \mathbf{I}_N - \mathbf{A} + \frac{1}{N} \mathbf{11^\top} \succeq \epsilon \mathbf{I}_N, \\
& A_{ii} = 0, A_{ij} \geq 0, \mathbf{A1 = 1}, \mathbf{A^\top = A}.
\end{array}
\end{align}
where we used that $\mathbf{L} = \mathbf{I}_N - \mathbf{A}$.

\section{Numerical simulations}

\begin{figure}[t]
\begin{center}
\includegraphics[width = 1\columnwidth]{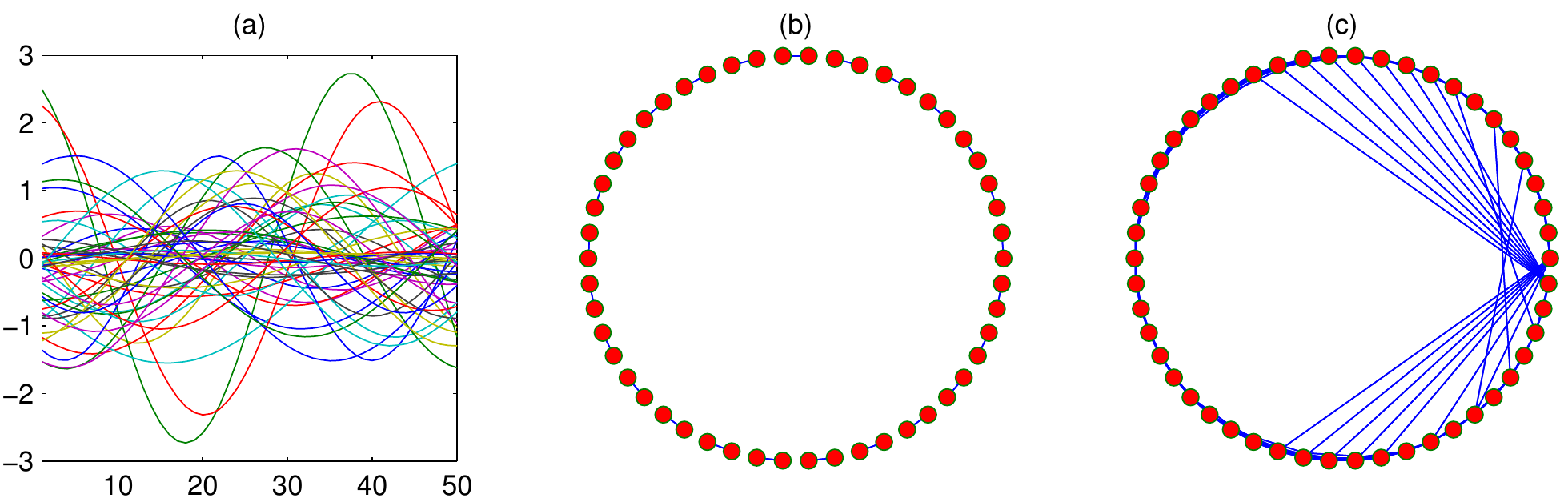}
\caption{Graph learning from data. (a) shows the sinusoids used to learn the graph, (b) shows reconstruction without the connectedness constraint and (c) shows reconstruction with the connectedness constraint. Edges with weights less than $0.001$ have been truncated.}
\vspace{-0.5cm}
\label{fig:sine_graph1}
\end{center}
\end{figure}

\begin{figure}[b]
\begin{center}
\vspace{-0.5cm}
\includegraphics[width = 0.9\columnwidth,height=0.4\columnwidth]{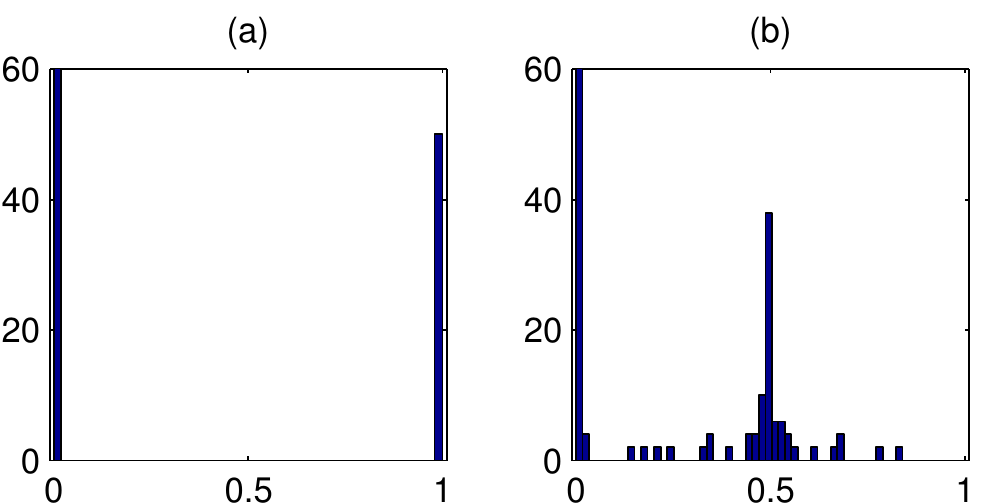}
\caption{Histogram of edge weights for learning graphs from data without the connectedness constraint (a) and with the constraint (b). The $y$-axis has been reduced to show the number of non-zero components more clearly.}
\vspace{-0.cm}
\label{fig:sine_graph1_histogram}
\end{center}
\end{figure}

We here show numerically that the constraint \eqref{cvx_constraint} ensures connectedness. We consider the problem of learning the weighted graph Laplacian from data using synthetic data and a data set of temperatures at Swedish cities \cite{venkitaraman2016temperature}. In the experiments we solved the optimization problem \eqref{eq:spg_learning_problem} using the CVX toolbox \cite{cvx} with $\epsilon = 0.01$.

\subsection{Experiments using synthetic data}

In the first experiment we synthetically generated the signals $\mathbf{X}\in\mathbb{R}^{M \times N}$ as
% sinusoids with random phases and frequencies as
\begin{align*}
X_{kl} = d_k \sin (f_k \cdot l + \phi_k),
\end{align*}
where $d_k \sim \mathcal{N}(0,1)$, $f_k \sim \mathcal{U}\left[\frac{2\pi}{N}-0.05,\frac{2\pi}{N}+0.05\right]$, $\phi_k \sim \mathcal{U}[-\pi,\pi]$ and $\mathcal{U}[a,b]$ denotes the uniform distribution on an interval $[a,b]$. In the experiment we set $N=M=50$. The data is shown in Figure~\ref{fig:sine_graph1} (a). We construct a graph describing the smoothness and periodicity of the data \cite{shuman2013emerging,7532684,chen2016representations} with and without the connectedness constraint. In Figure~\ref{fig:sine_graph1} (b) and (c) we see that without the connectedness constraint the graph shows local smoothness of the signals and is disconnected while the graph with the constraint shows local smoothness as well as some periodicity. In Figure~\ref{fig:sine_graph1}, the edges with weight less than $0.001$ were truncated. To examine if the graph is sensitive to the truncation we show the histogram of the adjacency matrices in Figure~\ref{fig:sine_graph1_histogram}. We find that without the constraint, only a few edges have large weights while the remaining are small. With the constraint, the weights assume a wider range of values.%more values in the range $0$ to $1$.

\begin{figure}[t]
\begin{center}
\includegraphics[width = 0.8\columnwidth,height=0.8\columnwidth]{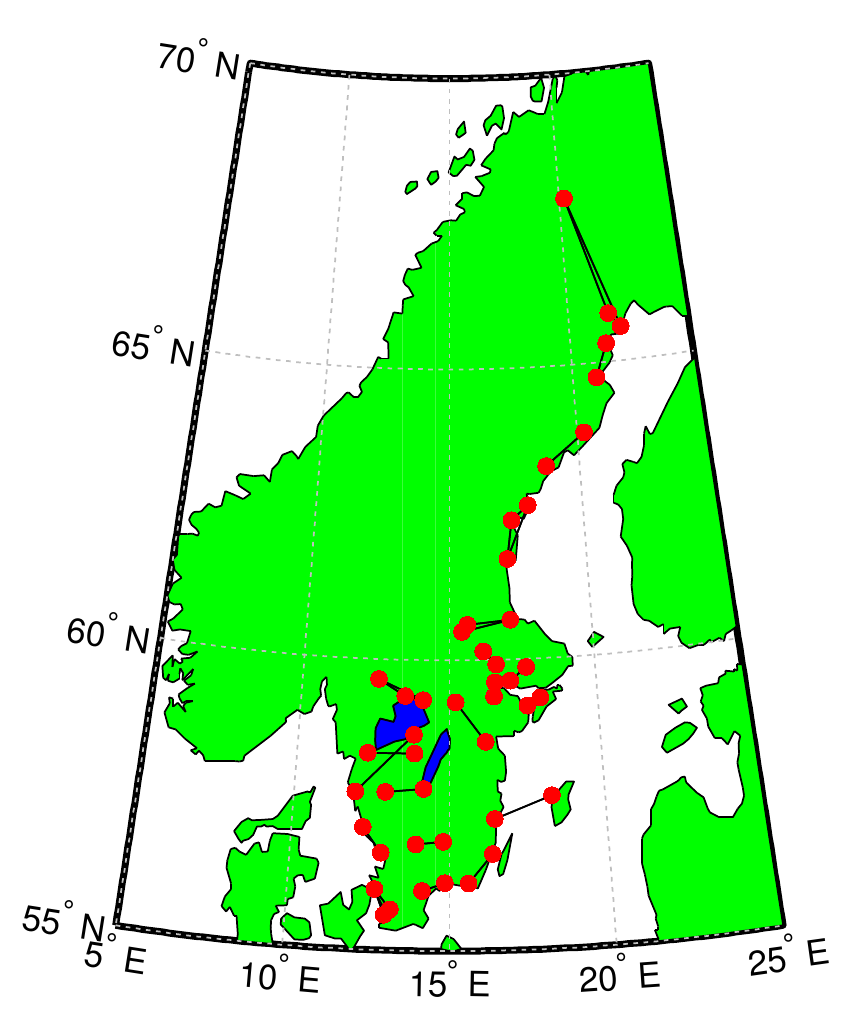}
\caption{Learned graph for the Swedish temperature dataset without the connectedness constraint.}
\vspace{-0.2cm}
\label{fig:sweden_map_noconstraint}
\end{center}
\end{figure}

\begin{figure}[t]
\begin{center}
\includegraphics[width = 0.8\columnwidth,height=0.8\columnwidth]{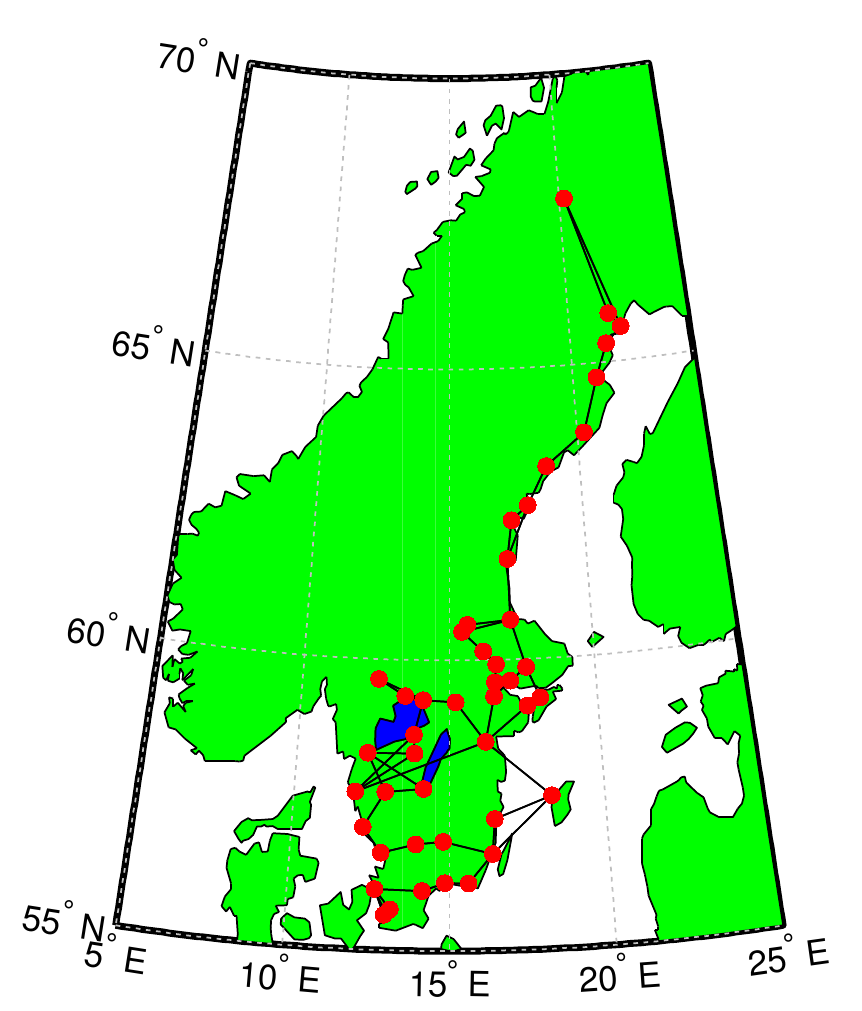}
\caption{Learned graph for the Swedish temperature dataset with the connectedness constraint.}
\vspace{-0.2cm}
\label{fig:sweden_map}
\end{center}
\end{figure}

\subsection{Experiments using real data}

\begin{figure}[]
\begin{center}
\includegraphics[width = 0.9\columnwidth,height=0.4\columnwidth]{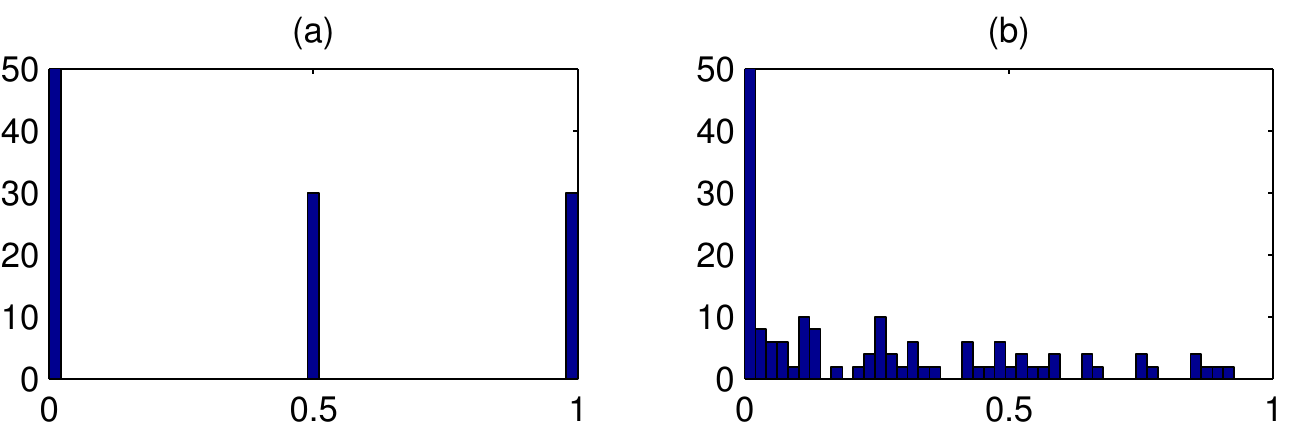}
\caption{Histogram of edge weights for learning graphs from temperature data without the connectedness constraint (a) and with the constraint (b). The $y$-axis has been reduced to show the number of non-zero components more clearly.}
\vspace{-0.5cm}
\label{fig:map_weights_histogram}
\end{center}
\end{figure}

In this experiment, we use time series of daily temperature data from 45 swedish cities from October to December of 2014 \cite{venkitaraman2016temperature}. Our task is to find a graph that shows which cities have similar temperatures. Note that in the experiment, the algorithm only has access to the temperatures and not the locations of the cities. We truncated edges with weights less than $0.05$. In Figure~\ref{fig:sweden_map_noconstraint} we see that the graph learned without the connectedness constraint does identify some neighboring cities but is disconnected while the graph with the constraint in Figure~\ref{fig:sweden_map} does identify neighboring cities from the temperature data. The graph also shows that cities on the same latitude have related temperatures. We show the histogram of edge weights in Figure~\ref{fig:map_weights_histogram}. We find that the edge weights have a wider range of values when the constraint is enforced.

%Greedy gLASSO on Minnesota dataset?
%
%Some weather data for signal processing on graphs?

\section{Conclusion}
\label{sec:concl}

In many problems, it is useful to represent relations between data using graphs. Often, the graph is not given a-priori but needs to be learned from data. Since graphs are often preferred to be sparse, it is a challenge to preserve connectedness of the graph while simultaneously enforcing sparsity. In this paper, we showed that connectedness is an analytical property that it can be imposed on a graph as a convex constraint making it possible to guarantee connectedness when learning sparse graphs. 
For the consensus problem, we showed that the graph is guaranteed to be disconnected for certain values of the regularization parameter when no constraint is imposed. We illustrated the effect of the constraint when learning a graph for synthetic data and for temperature data.

\section*{Acknowledgement}
This work was partially supported by the Swedish Research Council under contract 2015-05484. We also want to thank the anonymous reviewers for providing useful comments.

% References should be produced using the bibtex program from suitable
% BiBTeX files (here: strings, refs, manuals). The IEEEbib.bst bibliography
% style file from IEEE produces unsorted bibliography list.
% -------------------------------------------------------------------------

\bibliographystyle{IEEEbib}
\bibliography{strings,connected_graph_learning_arxiv}

\end{document}